\documentclass[final]{templ}

\usepackage{times}
\usepackage{epsfig}
\usepackage{graphicx}
\usepackage{amsfonts}   
\usepackage{amsmath}
\usepackage{amssymb}
\usepackage{multirow}
\usepackage{pifont}
\usepackage{xcolor}
\newcommand{\cmark}{\ding{51}}%
\newcommand{\xmark}{\ding{55}}
\usepackage[T1]{fontenc}
\usepackage{textcomp}
\usepackage{booktabs} 
\usepackage{caption}
\usepackage{subcaption}

\newcommand\blfootnote[1]{%
  \begingroup
  \renewcommand\thefootnote{}\footnote{#1}%
  \addtocounter{footnote}{-1}%
  \endgroup
}

\usepackage[pagebackref=true,breaklinks=true,colorlinks,bookmarks=false]{hyperref}

\begin{document}

\title{Rectification-based Knowledge Retention for Continual Learning}

\author{
Pravendra Singh$^{1,\ast}$, \hspace{0.5cm} Pratik Mazumder$^{2,\ast}$, \hspace{0.5cm} Piyush Rai$^2$, \hspace{0.5cm} Vinay P. Namboodiri$^{2,3}$ \\
$^1$Independent Researcher, India\hspace{0.5cm}
$^2$IIT Kanpur, India\hspace{0.5cm}
$^3$University of Bath, United Kingdom\\
{\tt\small pravendra1988@gmail.com, pratikm@cse.iitk.ac.in, piyush@cse.iitk.ac.in, vpn22@bath.ac.uk}
}
\maketitle

\begin{abstract}
Deep learning models suffer from catastrophic forgetting when trained in an incremental learning setting. In this work, we propose a novel approach to address the task incremental learning problem, which involves training a model on new tasks that arrive in an incremental manner. The task incremental learning problem becomes even more challenging when the test set contains classes that are not part of the train set, i.e., a task incremental generalized zero-shot learning problem. Our approach can be used in both the zero-shot and non zero-shot task incremental learning settings. Our proposed method uses weight rectifications and affine transformations in order to adapt the model to different tasks that arrive sequentially. Specifically, we adapt the network weights to work for new tasks by ``rectifying'' the weights learned from the previous task. We learn these weight rectifications using very few parameters. We additionally learn affine transformations on the outputs generated by the network in order to better adapt them for the new task. We perform experiments on several datasets in both zero-shot and non zero-shot task incremental learning settings and empirically show that our approach achieves state-of-the-art results. Specifically, our approach outperforms the state-of-the-art non zero-shot task incremental learning method by over 5\% on the CIFAR-100 dataset. Our approach also significantly outperforms the state-of-the-art task incremental generalized zero-shot learning method by absolute margins of 6.91\% and 6.33\% for the AWA1 and CUB datasets, respectively. We validate our approach using various ablation studies.
\end{abstract}

\section{Introduction}
\blfootnote{$^\ast$ The first two authors contributed equally.}
Deep learning models are used to solve many real-world problems, and they have even surpassed human-level performance in many tasks. However, deep learning models generally require all the training data to be available at the beginning of the training. If this is not the case, deep learning models suffer from catastrophic forgetting \cite{mccloskey1989catastrophic}, and their performance on the previously seen classes or tasks starts degrading. In contrast, human beings can continually learn new classes of data without losing the previously gained knowledge. To avoid catastrophic forgetting, deep learning models should perform lifelong/continual/incremental learning \cite{de2019continual,Parisi2019}. Deep learning models also require all the classes to be present in the train set. When the test set contains classes not seen during training, the performance of these models degrades significantly \cite{socher2013zero, zhang2017learning}. This is known as the zero-shot learning problem. The incremental learning problem becomes even more challenging when coupled with the zero-shot learning problem. In this paper, we solve for the task incremental learning problem in the zero-shot and non zero-shot setting.

The task incremental learning problem involves training the model on one task at a time, where each task has a set of non-overlapping classes. When a new task becomes available to the network for training, the previous task data is no longer accessible. Training on only the new task data causes the network to forget all the previous task knowledge. Therefore, the model has to prevent the forgetting of older tasks when training on new tasks. The task incremental generalized zero-shot learning problem involves training the model on one task at a time, where each task has a set of unseen classes not seen during training, which is the same as the zero-shot learning setting. The objective of the model is to successfully identify the seen and unseen classes of all the trained tasks. 

We propose a novel approach called Rectification-based Knowledge Retention (RKR) for the task incremental learning problem in the zero-shot and non zero-shot setting. Our approach (RKR) learns weight rectifications to adapt the network weights for a new task. After learning these weight rectifications, we can quickly adapt the network to work for images from that task by simply applying these weight rectifications to the network weights. We utilize an efficient technique for learning these weight rectifications to limit the model size. We also learn affine transformations (scaling factors) for all the intermediate outputs of the network that allow better adaptation of the network to the respective task.

We perform various experiments for the task incremental learning problem in both the zero-shot and non zero-shot settings in order to show the effectiveness of our approach. Using various ablation experiments, we validate the components of our approach. Our contributions can be summarized as follows:

\begin{itemize}
    \item We propose a novel approach for the task incremental learning problem in the zero-shot and non zero-shot settings that learns weight rectifications and scaling factors in order to quickly adapt the network to the respective tasks.
    \item RKR introduces very few parameters during training for learning the weight rectifications and scaling factors. The model size growth in our method is significantly low as compared to other dynamic network-based task incremental learning methods.
    \item We experimentally show that our method Rectification-based Knowledge Retention (RKR) significantly outperforms the existing state-of-the-art methods for the task incremental learning problem in both the zero-shot and non zero-shot settings. 
\end{itemize}

\section{Related Work}
\subsection{Incremental Learning}
Incremental learning is a setting where we have to train the model on tasks that arrive incrementally. The model has to retain the knowledge gained from the older task while learning the new tasks \cite{rebuffi2017icarl,aljundi2019task,yu2020semantic}. We can categorize incremental learning methods into: replay-based, regularization-based, and dynamic network-based methods. 

Replay-based methods require saving data from old tasks in order to fine-tune the network along with the new task data to reduce forgetting. The authors in \cite{rebuffi2017icarl} propose to use an exemplar-based prototype rehearsal technique along with distillation. The methods proposed in \cite{kemker2018fearnet,kamra2017deep} use a custom architecture to produce pseudo samples for the older tasks to be used for rehearsal. 

Regularization-based methods make use of regularization techniques to prevent network outputs from changing significantly while training on new tasks to preserve the knowledge gained from the older tasks. The work in \cite{li2017learning} proposes to use knowledge distillation as the regularization technique. In \cite{rebuffi2017icarl, castro2018end}, the authors propose to use modified classification techniques suited to continual learning in addition to the distillation loss. 

Dynamic network-based methods \cite{xiao2014error,fernando2017pathnet,chaudhry2018efficient,Oswald2020Continual,li2019learn} use network expansions/modifications for training new tasks. The work in \cite{rusu2016progressive} proposes to create an extra network for each new task with lateral connections to the networks of the older tasks. The method proposed in \cite{xu2018reinforced} uses reinforcement learning to determine how many neurons to add for each new task. DEN \cite{yoon2018lifelong} performs efficient selective retraining and dynamically expands the network for each task with only the required amount of units. The method proposed in \cite{rajasegaran2019random} uses a random path selection methodology for each task. The authors in \cite{Yoon2020Scalable} propose an order-robust approach APD, which uses task-shared and task-adaptive parameters. Recently the authors in \cite{singh2020ccll} proposed CCLL that calibrates the feature maps of convolutional layer outputs to perform incremental learning.

Our method RKR follows the dynamic network approach, but it is the first work that learns rectifications for the layer weights and outputs to adapt the model to any task quickly. Even though our method is dynamic network-based, it does not use the parameter isolation approach, which incrementally reserves a set of model parameters for new tasks. Therefore, our model will not run out of model capacity to accommodate future tasks. Our method introduces a significantly less number of parameters to learn the weight rectifications and scaling factors, e.g., for CIFAR-100 tasks using the ResNet-18 architecture, RKR introduces only 0.5\% additional parameters per task.

\subsection{Zero-Shot Learning} 

Zero-Shot Learning (ZSL) \cite{socher2013zero, zhang2017learning, zhao2018msplit, chen2018zero} is a problem setting where the model has to recognize classes not seen during training. In the generalized zero-shot learning setting, the test data can be from both the seen and unseen classes. Zero-shot learning methods utilize side information in the form of class embeddings/attributes that encode the semantic relationship between classes. The most popular way to deal with ZSL is to learn an embedding space where the image data and the class embedding are close to each other \cite{chen2018zero}. Another popular approach is to generate images/features of unseen classes by using their class embeddings \cite{felix2018multi, zhu2018generative}.  The authors in \cite{xian2018feature} propose f-CLSWGAN, which uses conditional Wasserstein GANs, to generate features for unseen classes. Cycle-WGAN \cite{felix2018multi} improves upon f-CLSWGAN by using reconstruction regularization in order to preserve the discriminative features of classes. CADA-VAE \cite{schonfeld2019generalized} uses a VAE to learn to generate class embedding features in a latent embedding space to train a zero-shot classifier. However, these methods require data replay to work in the continual learning setting \cite{wei2020lzsl}.

Recently, the authors in \cite{wei2020lzsl} proposed an approach LZSL for the task incremental generalized zero-shot learning problem. LZSL performs selective parameter retraining and knowledge distillation to preserve old domain knowledge and prevent catastrophic forgetting in the image feature encoder. We extend our proposed method RKR to the task incremental generalized zero-shot learning setting \cite{wei2020lzsl}. In this setting, RKR ``rectifies'' the weights and outputs of the image features encoder network in order to perform continual learning and prevent catastrophic forgetting.

\begin{figure*}[t]
    \centering
     \includegraphics[width=0.85\textwidth]{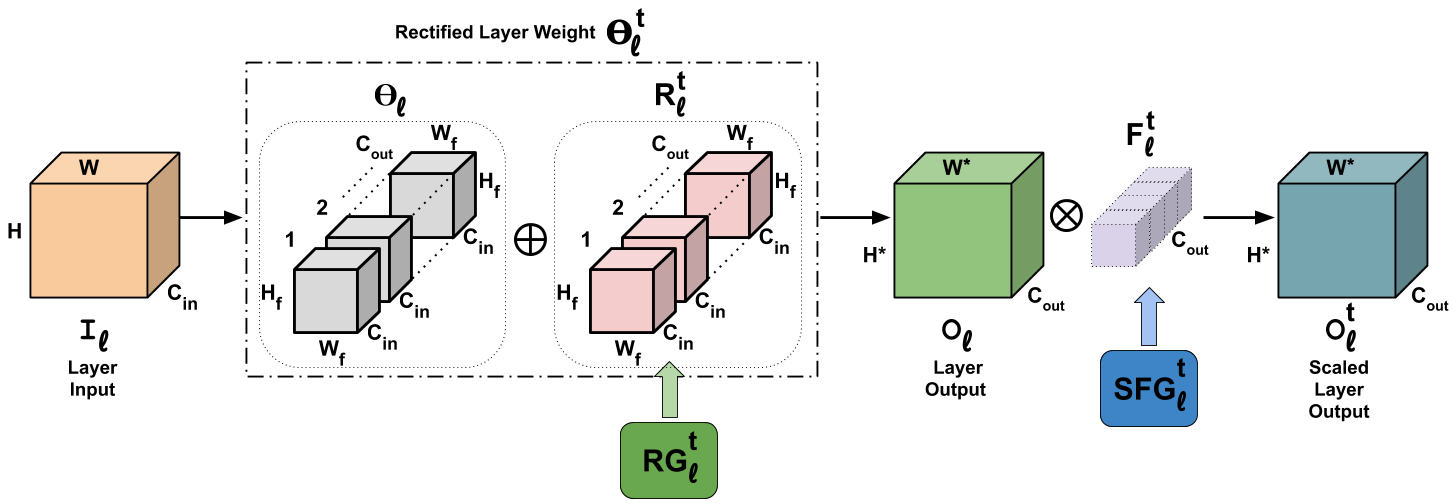}

    \caption{RKR for a convolutional layer. The weight rectifications $R_l^t$ produced by the rectification generator (RG$_l^t$) are added to the layer weights ($\Theta_l$) of the convolutional layer $l$ for task $t$. The task-adapted convolution layer weights ($\Theta_l^t$) are applied to the input $I_l$ to produce the layer output $O_l$. The scaling factor generator (SFG$_l^t$) produces scaling factors $F_l^t$ that are applied to $O_l$ to produce the scaled layer output ($O_l^t$). }

    \label{fig:rkrconv}
\end{figure*}

\section{Problem Definition}
\subsection{Task Incremental Learning}
In the task incremental learning setting, the network receives a sequence of tasks containing new sets of classes. When a new task becomes available, the previous task data are not accessible. The objective of task incremental learning is to obtain a model that performs well on the current task as well as the previous tasks.
\subsection{Task Incremental Generalized Zero-Shot Learning}
The task incremental generalized zero-shot learning setting also involves training the model on a sequence of tasks, but each task contains a set of seen and unseen classes, and the final model should perform well on the seen and unseen classes of the current task as well as the previous tasks. For this problem, we follow the setting defined in \cite{wei2020lzsl}, where each task is a separate dataset. When a new task becomes available for training, the older tasks are no longer accessible for further training/fine-tuning. 

\section{Proposed Method}
\subsection{Rectification-based Knowledge Retention}\label{sec:rkr}
We propose a task incremental learning approach called Rectification-based Knowledge Retention (RKR) that applies network weight rectifications and scaling transformations to adapt the network to different tasks.

Let us assume that we have a deep neural network with $N$ layers, i.e., \{$L_1, L_2..., L_{N}$\}. Each layer can be a convolutional layer or a fully connected layer. Let $\Theta_l$ represent the parameter weights of layer $L_l$.  If we train this network on a task containing a set of classes, the network will learn the parameter weights $\Theta_l$ for each layer $l\in \{1, 2,.., N\}$. However, if we then train the network on a new task (with a new set of classes), it will learn new parameter weights $\Theta_l^{\ast}$ to work for this task and will lose information regarding the previous tasks (catastrophic forgetting).

\begin{figure}[t]
    \centering
     \includegraphics[width=0.45\textwidth]{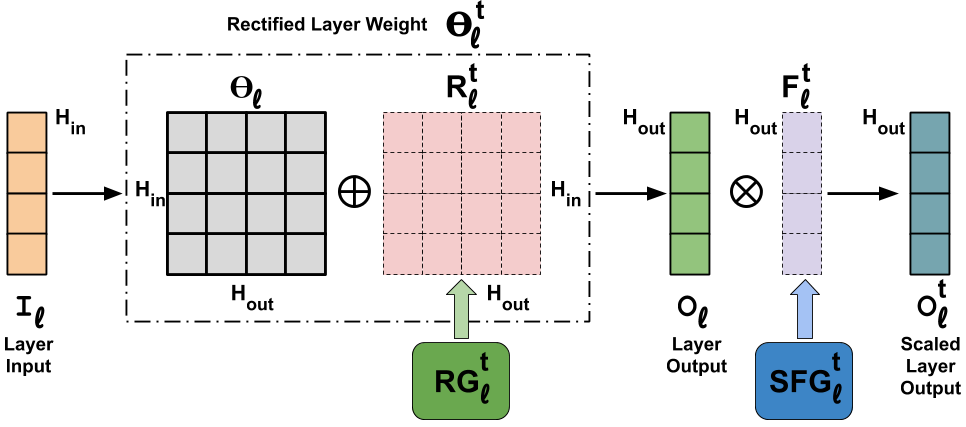}

    \caption{RKR for a fully connected layer. The weight rectifications $R_l^t$ produced by the rectification generator (RG$_l^t$) are added to the layer weights ($\Theta_l$) of the fully connected layer $l$ for task $t$. The task-adapted fully connected layer weights ($\Theta_l^t$) are applied to the input $I_l$ to produce the layer output $O_l$. The scaling factor generator (SFG$_l^t$) produces scaling factors $F_l^t$ that are applied to $O_l$ to produce the scaled layer output ($O_l^t$).}

    \label{fig:rkrfc}
    
\end{figure}

We propose to avoid this problem using the dynamic network-based approach. For each task, we learn the rectifications needed to adapt the layer weights of the network to work for that task. Let $R_l^t$ refer to the weight rectification needed to adapt the $l^{th}$ layer of the network to work for task $t$. We use a rectification generator (RG) for learning these rectifications. RG uses very few parameters to learn the weight rectifications as described in Sec. \ref{sec:paramred}. The weight rectifications $R_l^t$ are added to the layer weights $\Theta_l$ for each task $t$ (Figs. \ref{fig:rkrconv}, \ref{fig:rkrfc}). 
\begin{equation}\label{eq:wtcor}
    \Theta_l^{t} = \Theta_l \oplus R_l^t 
\end{equation}
where $\Theta_l$ refers to weights of layer $l$ of the network, $R_l^t$ refers to the rectifications to be learned for the weights of the layer $l$ for task $t$, $\Theta_l^{t}$ refers to the rectified weights of the layer $l$ for task $t$, $\oplus$ refers to element-wise addition. The layer weight $\Theta_l$ is trained only on the first task and is adapted using the weight rectifications $R_l^t$ (that are learned for all tasks) to obtain $\Theta_l^t$.

Apart from the weight rectifications, we also learn scaling factors for performing affine transformations on the intermediate outputs generated by each layer of the network. We use a scaling factor generator (SFG) for learning the scaling factors. In the case of a fully connected layer $l$, the scaling factors $F_l^t$ have the same size as the layer output $O_l$, and we multiply them element-wise to each component of $O_l$ (Fig.~\ref{fig:rkrfc}). In the case of a convolutional layer $l$, the scaling factors $F_l^t$ have the same number of elements as the number of feature maps in $O_l$, and we multiply them to the corresponding feature maps of $O_l$ (Fig.~\ref{fig:rkrconv}). These learned scaling factors represent the rectifications needed to adapt the intermediate network outputs to the corresponding task.
\begin{equation}
    O_l^{t} = O_l \otimes F_l^t
\end{equation}
where $O_l$ refers to the output from the layer $l$ of the network, $F_l^t$ refers to the scaling factors learned for the output of the layer $l$ of the network for task $t$, $\otimes$ refers to the scaling operation, and $O_l^{t}$ denotes the scaled layer output for task $t$. Our approach applies the weight rectifications and scaling factors to quickly adapt the network for any task $t$.

\subsection{Reducing Parameters for Weight Rectification}\label{sec:paramred}

\subsubsection{Convolutional Layers}
The weight rectifications for a convolutional layer requires to be of the same size as the convolutional layer weights. Let $W_{f}$,$H_{f}$,$C_{in}$ be the width, height and number of channels of each filter of the convolutional layer $L_{l}$ and $C_{out}$ be the number of filters used in $L_{l}$. Therefore, the total size of the convolutional layer weights is $W_f\times H_f\times C_{in}\times C_{out}$. The weight rectification $R_l^t$ for layer $L_{l}$ has to be of the same size. In order to reduce the number of parameters needed to generate these weight rectifications, we use a rectification generator (RG). The rectification generator (RG) learns two matrices of smaller size i.e., $LM_l^t$ of size $(W_f\ast C_{in})\times K$ and $RM_l^t$ of size $K\times (H_f\ast C_{out})$, where $\ast$ represents scalar multiplication. Here, $K \ll (W_f\ast C_{in})$ and $K \ll (H_f\ast C_{out})$. This matrix factorization ensures that we introduce very few parameters to generate these weight rectifications. The product of these two matrices produces the weight rectifications which are reshaped to the size $W_f\times H_f\times C_{in}\times C_{out}$ and added to the convolutional layer weights element-wise. Therefore, RG computes the weight rectifications $R_l^t$ for task $t$ as:
\begin{equation}
    R_l^t = \texttt{MATMUL}(LM_l^t,RM_l^t)
\end{equation}
where $\texttt{MATMUL}$ refers to matrix multiplication.

We apply the adapted convolution layer weights ($\Theta_l^t$) to the input ($I_l$) of size, say, $W\times H\times C_{in}$, to obtain an output of size $W'\times H'\times C_{out}$ (See Fig. \ref{fig:rkrconv}). Here, $W$, $W'$ refer to the width of the feature maps before and after applying the convolution. $H$, $H'$ refer to the height of the feature maps before and after applying the convolution. We then apply scaling transformation to the output $O_l$. The scaling factor generator (SFG) learns a scaling parameter for each feature map. Therefore, SFG introduces a very insignificant number of parameters, i.e., $C_{out}$, which is equal to the number of feature maps in $O_l$. 
\begin{equation}
    EP_{conv}=\frac{K\ast(W_f\ast C_{in} + H_f\ast C_{out}) + C_{out}}{W_f\ast H_f\ast C_{in}\ast C_{out}}\ast 100
\end{equation}
where $EP_{conv}$ refers to the percentage of extra parameters introduced by our approach for each convolutional layer.

\subsubsection{Fully Connected Layers}

The weight rectifications $R_l^t$ for the fully connected layers require to be of the same size as the layer weights $\Theta_l$. Let $\Theta_l$ be of size $H_{in}\times H_{out}$. Here $H_{in}$ and $H_{out}$ refer to the size of input and output of the fully connected layer, respectively. In order to reduce the number of parameters needed to generate the weight rectifications, the rectification generator (RG) learns two matrices of smaller size, i.e., $LM_l^t$ of size $H_{in}\times K$ and $RM_l^t$ of size $K\times H_{out}$. Here, $K < H_{in}$ and $K < H_{out}$. Therefore, the total parameters added will not be significant. The product of these two matrices will give the weight rectifications of size $H_{in}\times H_{out}$ which we add to the layer weights $\Theta_l$ element-wise to produce the adapted layer weight $\Theta_l^t$. Therefore, RG computes the weight rectifications $R_l^t$ for task $t$ as follows:
\begin{equation}
    R_l^t = \texttt{MATMUL}(LM_l^t,RM_l^t)
\end{equation}
where $\texttt{MATMUL}$ refers to matrix multiplication.

We apply the adapted fully connected layer weights ($\Theta_l^t$) to the input ($I_l$) of size $H_{in}$, to obtain an output ($O_l$) of size $H_{out}$ (See Fig. \ref{fig:rkrfc}). We then apply scaling transformation to the output $O_l$. The scaling factor generator (SFG) learns a parameter for each component of $O_l$. Therefore, SFG introduces a very insignificant number of parameters, i.e., $H_{out}$.

\begin{equation}
    EP_{fc}=\frac{K\ast(H_{in} + H_{out}) + H_{out}}{H_{in}\ast H_{out}}\ast 100
\end{equation}
where, $EP_{fc}$ refers to the percentage of extra parameters introduced by our approach for each fully connected layer.

Therefore, our approach introduces very few parameters per task to learn weight rectifications, and scaling factors in the incremental learning setting, e.g., for the ResNet-18 architecture RKR introduces only 0.5\% additional parameters per ImageNet-1K task. Intuitively, this simulates separate networks for each task using very few parameters, e.g., our model for ImageNet-1K with 10 tasks has (100 $+$ 10 $\ast$ 0.5)\% capacity. However, directly using separate networks will lead to an impractical model with (100 $\ast$ 10)\% capacity.

\section{Task Incremental Learning Experiments (Non Zero-shot Setting)}

\subsection{Datasets}
We perform the task incremental learning experiments on the CIFAR \cite{krizhevsky2009learning} and ImageNet \cite{russakovsky2015imagenet} datasets for the non zero-shot setting. We perform experiments on CIFAR-100 with 10 tasks where each task contains 10 classes. For split CIFAR-10/100 experiments, we use all the classes of CIFAR-10 for the first task and randomly choose 5 tasks of 10 classes each from CIFAR-100. So we have 6 tasks for this setting. In the case of ImageNet-1K, we group the 1000 classes into 10 tasks of 100 classes each. 

\subsection{Implementation Details}

In our approach, we learn weight rectifications and scaling factors for each convolutional layer and fully connected layer of the network (except the classification layer). We train the full network on the first task (base network). For every new task, we only learn weight rectifications and scaling factors for all network layers to adapt them to the new task. During testing, depending on the task, we apply the corresponding weight rectifications and scaling factors to the base network to work for that task.

For CIFAR-100 experiments, we use the ResNet-18 architecture \cite{he2016deep}. For split CIFAR-10/100 experiments, we use ResNet-32 architecture \cite{he2016deep}. In the above experiments, we train the network for 150 epochs for each task with the initial learning rate as 0.01, and we multiply the learning rate by 0.1 at the 50, 100, and 125 epochs. We also perform experiments with the LeNet architecture \cite{lecun1998gradient} on CIFAR-100. We train the network on each task for 100 epochs with the initial learning rate as 0.01 and multiply the learning rate with 0.5 at the 20, 40, 60, and 80 epochs. For ImageNet-1K experiments, we use the ResNet-18 architecture and train the network for 70 epochs for each task with the initial learning rate as 0.01, and we multiply the learning rate by 0.2 at the 20, 40, and 60 epochs. We use the SGD optimizer in all our experiments. In all the cases, we run experiments for 5 random task orders and report the average accuracy. We perform experiments with $K=2$ since this is a good choice considering the accuracy/extra-parameters trade-off as shown in Table~\ref{tab:cifar_abl_k}. Our method utilizes task labels during testing similar to \cite{singh2020ccll,Yoon2020Scalable,yoon2018lifelong}.

\subsection{CIFAR-100 Results} 

\begin{figure}[t]
    \centering
     \includegraphics[width=0.4\textwidth]{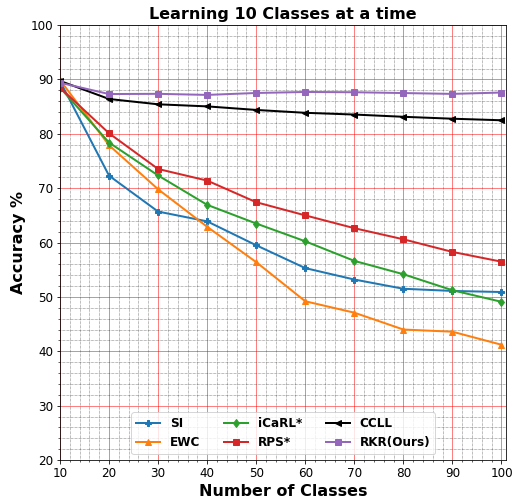}

    \caption{Experimental results on CIFAR-100 using ResNet-18. `$*$' denotes replay-based approach.}\label{fig:result_cifar100}

\end{figure}

For the experiments on the CIFAR-100 incremental learning tasks using 10 classes at a time, we perform experiments with various methods such as CCLL \cite{singh2020ccll}, SI \cite{zenke2017continual}, EWC \cite{kirkpatrick2017overcoming}, iCARL \cite{rebuffi2017icarl} and RPS \cite{rajasegaran2019random}. CCLL uses task labels at test time, and we modify SI and EWC to use task labels during testing for a fair comparison. We additionally report the result for iCaRL, RPS-Net which are replay-based methods. These methods store data from the previous tasks as additional data to use this data along with the new task data to train the network to reduce catastrophic forgetting. RKR does not store previous task data and is more scalable in this regard. We provide these additional results for completeness (replay-based and non replay-based). However, replay-based should not be directly compared with RKR. Figure \ref{fig:result_cifar100} indicates that our approach RKR outperforms all existing methods. RKR outperforms CCLL \cite{singh2020ccll} by an absolute margin of 5.1\% in the overall accuracy. Our approach performs consistently better than all other methods as more tasks arrive. 

RKR applies the weight corrections to adapt the network to the new task, which is very natural and intuitive because training on a new task changes the network layer weights and, consequently, the corresponding features. In contrast, CCLL adapts the network to the new task by only calibrating the convolutional layer output feature maps that are biased to the initial task. This is the reason why we see a significant performance gap between RKR and CCLL. This problem becomes even more apparent if the new task is very different from the initial task. In such a case, the features extracted by the model trained on the initial task will not be relevant to the new task, and it will be very difficult to calibrate the feature maps to correctly estimate the feature maps of the new task. For example, on taking MNIST images in the initial task and taking 10 tasks of CIFAR-100 as the subsequent tasks, the performance gap between RKR and CCLL increases from 5.1\% to 16\% absolute margin.

\begin{table}[t]
\begin{center}
            \scalebox{0.85}{
            \addtolength{\tabcolsep}{5pt}
            \begin{tabular}{lcc}
            \hline
          Methods & Capacity & Accuracy  \\
             \hline
            L2T \cite{Yoon2020Scalable}& 100\% & 48.73\%  \\
            EWC \cite{kirkpatrick2017overcoming}& 100\% & 53.72\%   \\
            P\&C \cite{schwarz2018progress} ICML'18 & 100\% & 53.54\%  \\
            PGN \cite{rusu2016progressive} & 171\% & 54.90\%   \\
            RCL \cite{xu2018reinforced} NIPS'18 & 181\% & 55.26\%  \\            
            DEN \cite{yoon2018lifelong} ICLR'18 & 181\% & 57.38\%  \\
             
            APD \cite{Yoon2020Scalable} ICLR'20 & 135\% & 60.74\% \\
            CCLL \cite{singh2020ccll} NIPS'20 & 100.7\%  & 63.71\% \\
            \hline
            RKR-Lite (Ours) & 100.7\% & \textbf{66.32}\% \\
            RKR (Ours) & 104.3\% & \textbf{69.58}\% \\
            \hline
            \end{tabular}}
        \caption{Experimental results on CIFAR-100 using LeNet.}
                    \label{tab:cifar_lenet}
            \end{center}
\end{table}

We also perform experiments on the CIFAR-100 tasks using the LeNet architecture (20-50-800-500) as used in \cite{Yoon2020Scalable}. All the methods compared in Table~\ref{tab:cifar_lenet} use task labels during testing. The results in Table~\ref{tab:cifar_lenet} indicate that RKR outperforms existing state-of-the-art methods. Specifically, our approach RKR outperforms CCLL by an absolute margin of 5.87\%. We also report the results for RKR-Lite, which uses only weight rectifications for the convolutional layers and only scaling factors for the fully connected layers. RKR-Lite introduces the same number of parameters as CCLL but outperforms it by an absolute margin of 2.61\%.

\begin{figure}[t]
    \centering
     \includegraphics[width=0.44\textwidth]{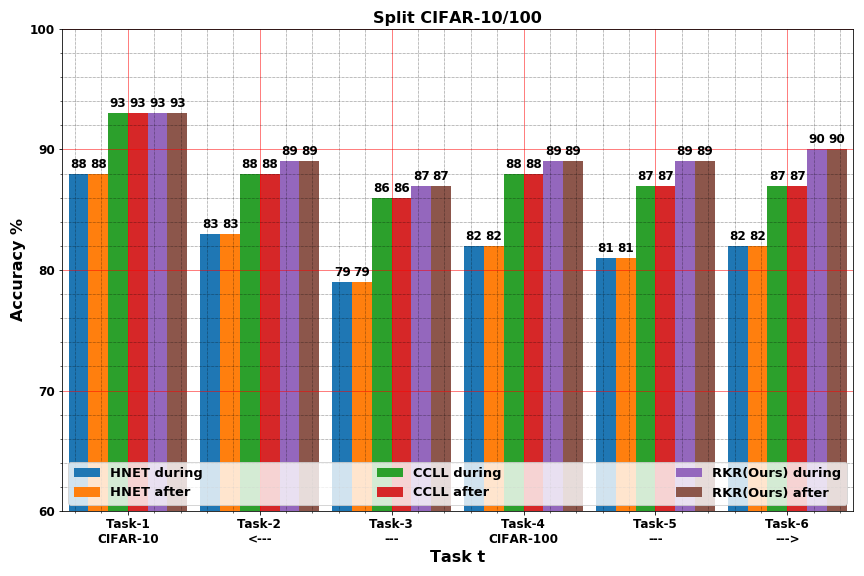}
    \caption{Experimental results on split CIFAR-10/100 using ResNet-32 to check for catastrophic forgetting. We report the achieved accuracy for each task when the network is trained on that task (marked as during) and after the network has been trained on all the tasks (marked as after).}
    \label{fig:result_cifar10-100_bargraph}
\end{figure}
\textbf{Split CIFAR-10/100}: For the split CIFAR-10/100 tasks we use the ResNet-32 architecture and compare our method RKR with CCLL and HNET \cite{Oswald2020Continual}, which are the state-of-the-art methods for this setup and use task labels during testing. We observe in Fig. \ref{fig:result_cifar10-100_bargraph} that RKR prevents catastrophic forgetting just like CCLL and HNET. Therefore, RKR helps in avoiding catastrophic forgetting without significantly affecting the network's ability to learn each task properly. 

\begin{figure}[t]
    \centering
     \includegraphics[width=0.4\textwidth]{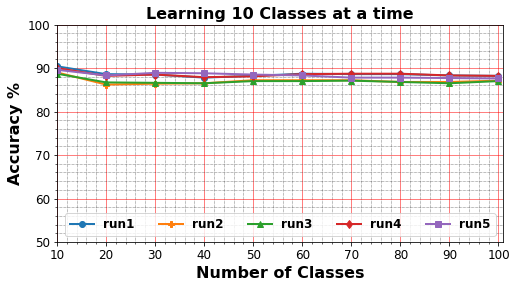}
    \caption{Experimental results for 5 runs of task incremental learning experiments using RKR on the CIFAR-100 dataset with ResNet-18.}

    \label{fig:run5}

\end{figure}

\textbf{Different First Task:} As mentioned in Sec.~\ref{sec:rkr}, $\Theta_l$ is trained only on the first task. Therefore, we perform experiments with different first tasks. From Fig.~\ref{fig:run5}, we observe that the performance of RKR is stable for different first tasks with different task orders.

\textbf{Value of K:} We perform experiments on the CIFAR-100 with ResNet-18 using different values of K. The results in Table \ref{tab:cifar_abl_k} indicates that for $K=2$ is a good  choice while considering extra-parameters/accuracy trade-off. Therefore, we use $K=2$ for all our experiments. We also observe that the FLOPS increase due to RKR is insignificant.

\begin{table}
\centering
\scalebox{0.8}{
\addtolength{\tabcolsep}{5pt}
\begin{tabular}{cccc}
\hline
K & \% Params. $\uparrow$ & \% FLOPS $\uparrow$ & Accuracy  \\
\hline
 1 & 0.3355\% & $8.6\times1e\texttt{-}4$\% &   85.55\% \\
 2 &  0.5426\% & $8.6\times1e\texttt{-}4$\% &   87.60\% \\
 4 &  0.9569\% & $8.6\times1e\texttt{-}4$\% &   87.90\% \\
 8 &  1.7854\% & $8.6\times1e\texttt{-}4$\% &   88.49\% \\
\hline
\end{tabular}}
\caption{Experimental results on CIFAR-100 with ResNet-18 using RKR for different values of K. We report the average accuracy of the 10 tasks.
}
\label{tab:cifar_abl_k}
\end{table} 

\begin{table}

\centering
\scalebox{0.8}{
\addtolength{\tabcolsep}{-2pt}
\begin{tabular}{lcccccc}
\hline
Arch. & Wt. Rect. & Scaling & \% Params. $\uparrow$ & \% FLOPS $\uparrow$ & Acc.  \\
\hline
LeNet  & \xmark & \cmark  & 0.0497\% & $6.4\times1e\texttt{-}4$\% &   62.3\% \\
  & \cmark & \xmark  & 0.3795\% & 0.0\% &   69.1\% \\  
  & \cmark & \cmark  & 0.4292\% & $6.4\times1e\texttt{-}4$\% &   69.6\% \\  
\hline
Res-18  & \xmark & \cmark  & 0.1283\% & $8.6\times1e\texttt{-}4$\% &   79.2\% \\
  & \cmark & \xmark  & 0.4143\% & 0.0\% &   87.5\% \\  
  & \cmark & \cmark  & 0.5426\% & $8.6\times1e\texttt{-}4$\% &   87.6\% \\  

\hline
\end{tabular}}
\caption{Experimental results on CIFAR-100 with LeNet and ResNet-18 using RKR ($K=2$) with different components. \cmark\ and \xmark\ refer to presence and absence respectively. 
}
\label{tab:cifar_abl_comp}
\end{table}

\begin{table*}[t]
\footnotesize

\begin{center}
\scalebox{0.9}{
\begin{tabular}{l c c c c c c c c c c}
\hline
 \textbf{Method} & \textbf{1} & \textbf{2}& \textbf{3}& \textbf{4}& \textbf{5}& \textbf{6}& \textbf{7}& \textbf{8}& \textbf{9}& \textbf{Final Acc.} \\
            
            \hline
 iCaRL$^\ast$ \cite{rebuffi2017icarl} CVPR'17 & 90.1 &  82.8 & 76.1 & 69.8 & 63.3 & 57.2 & 53.5 & 49.8 & 46.7  & 44.1 \\
 RPS-Net$^\ast$ \cite{rajasegaran2019random} NIPS'19 & 90.2 & 88.4 & 82.4 & 75.9 & 66.9 & 62.5 & 57.2 & 54.2 & 51.9 & 48.8 \\
 EEIL$^\ast$ \cite{castro2018end} ECCV'18& 95.0 & 95.5 & 86.0 & 77.5 & 71.0 & 68.0 & 62.0 & 59.8 & 55.0 & 52.0 \\
 CCLL~\cite{singh2020ccll} NIPS'20& 91.4 & 88.3 & 86.5 & 86.6 & 84.6 & 83.5 & 82.7 & 81.7 & 81.2 & 81.3 \\
 RKR(Ours) & 90.2 & 88.7 & 88.1 & 88.2 & 86.6 & 85.7 & 85.0 & 84.2 & 83.8 &  \ \ \ \ \ \ \ \ $\textbf{84.4}_{\textcolor{red}{+3.1}}$ \\

\hline
\end{tabular}
}

\caption{Task incremental learning experiments on the ImageNet-1K dataset with 10 tasks. The reported accuracy for each task is the average of all accuracies up to that task. `$*$' denotes replay-based approach.}
\label{tab:imagenet}
\end{center}

\end{table*}

 \begin{table*}[t]
\begin{center}
\scalebox{0.8}{

\begin{tabular}{ l | c | ccc | ccc | ccc | ccc }
\hline
Method  & Total & \multicolumn{3}{c}{aPY}  &  \multicolumn{3}{c}{AWA1}  & \multicolumn{3}{c}{CUB}  & \multicolumn{3}{c}{SUN} \\
  
\cline{3-14}
  & Mem. & U   & S   & H   & U   & S   & H
 & U   & S   & H  & U   & S   & H \\
 \hline
Base & 100\% &  6.69 & 0.59 & 1.09 & 5.14 & 0.92 & 1.56 & 0.87 & 0.67 & 0.76 & 43.40 & 33.95 & 38.10
\\
SFT & 100\%  &  24.24 & 23.21 & 23.71 & 47.27 & 55.18 & 50.92 & 35.46 & 34.74 & 35.10 & 38.47 & 36.10 & 37.20
\\
L1 & 200\%  &  26.42 & 29.79 & 28.01 & 49.64 & 58.23 & 53.59 & 35.11 & 32.31 & 33.65 & 40.14 & 34.11 & 36.88
\\
L2 & 200\%  &  24.08 & 23.61 & 23.84 & 46.71 & 59.07 & 52.17 & 35.53 & 33.24 & 34.35 & 42.08 & 32.33 & 36.56
\\
LZSL & 200\%  &  29.11 & 43.29 & 34.81 & 51.17 & 63.66  & 56.73 & 38.82 & 45.81 & 42.03 & 42.43 & 31.78 & 36.34
\\
\textbf{RKR}(Ours) & 113\% &  33.39 & 51.34 & \textbf{40.46} & 58.79 & 69.36 & \textbf{63.64} & 47.52 & 49.22 & \textbf{48.36} & 42.22 & 36.01 & \textbf{38.87}
\\
\hline
Original & 400\% &  30.36 & 59.36 & 40.18 & 57.30 & 72.80 & 64.10 & 53.50 & 51.60 & 52.40 & 35.70 & 47.20 & 42.60
\\
\hline
\end{tabular}
}
\end{center}

\caption{Classification accuracy (\%) of incremental generalized zero-shot learning on the sequence of datasets aPY, AWA1, CUB, and SUN for our method RKR and other methods. LZSL~\cite{wei2020lzsl} is the state-of-the-art method in this setting.}

\label{tab:mainres}
\end{table*}

\textbf{Significance of Components:} In Table \ref{tab:cifar_abl_comp}, we observe that without weight corrections, the model performs lower by absolute margins of 7.3\% and 8.4\% for CIFAR-100 using LeNet and ResNet-18, respectively. Without scaling, the model performance suffers slightly. However, we observe in task incremental generalized zero-shot learning that scaling helps to improve the RKR performance (Sec. \ref{sec:zerores}).

\textbf{Forward Knowledge Transfer:} In RKR, when we train the model on a new task, we initialize the parameters of RG and SFG from the previous task. If we train these parameters from scratch for every new task, the model performance falls by an absolute margin of $1.45\%$ for CIFAR-100 using the ResNet-18. Therefore, forward transfer of knowledge occurs in RKR.

\subsection{ImageNet-1K Results}
From Table~\ref{tab:imagenet}, we observe that RKR significantly outperforms the state-of-the-art CCLL method. Specifically, our method outperforms CCLL by an absolute margin of $3.1$\% (top-5 accuracy) even though both RKR and CCLL introduce around 0.5\% extra parameters per task. It should also be noted that CCLL introduces 0.98\% extra FLOPS in the model, whereas RKR introduces only $2.8\times1e\texttt{-}4$\% extra FLOPS, which is very insignificant.

\section{Task Incremental Learning Experiments (Generalized Zero-Shot Setting)}

\subsection{Datasets}

We experiment with four datasets for the task incremental generalized zero-shot learning (GZSL) problem i.e. Attribute Pascal and Yahoo (aPY) \cite{farhadi2009describing}, Animals with Attributes 1 (AWA1) \cite{xian2018zero}, Caltech-UCSD-Birds 200-2011 (CUB) \cite{wah2011caltech}, and SUN Attribute dataset (SUN) \cite{patterson2012sun}. We extract the image features of 2048 dimensions from the final pooling layer of an ImageNet pre-trained ResNet-101. We follow the training split proposed in \cite{xian2018zero} so that the test classes do not overlap with the training classes. 

\begin{table}[b]
\begin{center}
\resizebox{0.45\textwidth}{!}{%
\addtolength{\tabcolsep}{-4.4pt}
\begin{tabular}{ ccc | ccc | ccc | ccc }
\toprule
  \multicolumn{3}{c}{aPY}  &  \multicolumn{3}{c}{AWA1}  & \multicolumn{3}{c}{CUB}  & \multicolumn{3}{c}{SUN} \\
\hline
   U   & S   & H   & U   & S   & H
 & U   & S   & H  & U   & S   & H \\
 \hline
  33.39 & 51.34 & 40.46 & 58.79 & 69.36 & 63.64 & 47.52 & 49.22 & 48.36 & 42.22 & 36.01 & 38.87
\\

\toprule
  \multicolumn{3}{c}{AWA1}  &  \multicolumn{3}{c}{aPY}  & \multicolumn{3}{c}{CUB}  & \multicolumn{3}{c}{SUN} \\
  
\hline
   U   & S   & H   & U   & S   & H
 & U   & S   & H  & U   & S   & H \\

\hline
  61.93 & 66.49 & 64.13 & 30.96 & 55.25 & 39.68 & 48.06 & 50.36 & 49.18 & 47.08 & 31.78 & 37.95
\\

\toprule
  \multicolumn{3}{c}{CUB}  &  \multicolumn{3}{c}{AWA1}  & \multicolumn{3}{c}{aPY}  & \multicolumn{3}{c}{SUN} \\
  
\hline

    U   & S   & H   & U   & S   & H
 & U   & S   & H  & U   & S   & H \\
\hline

  51.11 & 53.88 & 52.46 & 56.02 & 70.01 & 62.24 & 30.82 & 53.39 & 39.08 & 46.25 & 32.05 & 37.87
\\

\toprule

   \multicolumn{3}{c}{SUN}  &  \multicolumn{3}{c}{AWA1}  & \multicolumn{3}{c}{CUB}  & \multicolumn{3}{c}{aPY} \\
  
\hline

    U   & S   & H   & U   & S   & H
 & U   & S   & H  & U   & S   & H \\
 \hline

  45.28 & 36.67 & 40.52 & 57.81 & 67.91 & 62.46 & 47.51 & 49.48 & 48.47 & 31.21 & 57.87 & 40.55
\\
\bottomrule

\end{tabular}
}
\end{center}

\caption{Experimental results for RKR with different first dataset in the task incremental GZSL problem.}
\label{tab:ablseq}

\end{table}

\begin{table*}[!t]
\begin{center}
\scalebox{0.8}{

\begin{tabular}{ c|c| ccc | ccc | ccc | ccc }
\hline
 Wt. & Scaling &  \multicolumn{3}{c}{aPY}  &  \multicolumn{3}{c}{AWA1}  & \multicolumn{3}{c}{CUB}  & \multicolumn{3}{c}{SUN} \\
  
\cline{3-14}
 Rec. &  & U   & S   & H   & U   & S   & H
 & U   & S   & H  & U   & S   & H \\
 \hline
 \cmark & \xmark & 32.34 & 52.78 & 40.10 & 52.50 & 69.80 & 59.93 & 45.12 & 42.54 & 43.79 & 40.97 & 30.97 & 35.28
\\
 \xmark & \cmark & 29.97 & 52.08 & 38.05 & 52.82 & 61.53 & 56.84 & 39.17 & 39.00 & 39.08 & 36.81 & 29.30 & 32.63
\\
 \cmark & \cmark  & 33.39 & 51.34 & 40.46 & 58.79 & 69.36 & 63.64 & 47.52 & 49.22 & 48.36 & 42.22 & 36.01 & 38.87
\\

\hline
\end{tabular}
}
\end{center}

\caption{Classification accuracy (\%) of task incremental generalized zero-shot learning on the aPY, AWA1, CUB and SUN datasets using RKR with different combinations of its components. Wt. Rec. refers to weight rectifications.}

\label{tab:ablcomp}
\end{table*}

\begin{table*}[t]
\begin{center}
\scalebox{0.8}{

\begin{tabular}{ c|c| ccc | ccc | ccc | ccc }
\hline
 K & Train &  \multicolumn{3}{c}{aPY}  &  \multicolumn{3}{c}{AWA1}  & \multicolumn{3}{c}{CUB}  & \multicolumn{3}{c}{SUN} \\
  
\cline{3-14}
  & Mem. & U   & S   & H   & U   & S   & H
 & U   & S   & H  & U   & S   & H \\
 \hline
 1 & 101\% & 30.33 & 58.78 & 40.01 & 55.66 & 69.49 & 61.81 & 40.11 & 40.68 & 40.39 & 40.56 & 30.50 & 34.82
\\
 4 & 104\% & 31.10 & 56.55 & 40.13 & 55.44 & 69.16 & 61.55 & 39.37 & 47.95 & 43.24 & 42.71 & 30.78 & 35.77
\\
 16 & 113\%  & 33.39 & 51.34 & 40.46 & 58.79 & 69.36 & 63.64 & 47.52 & 49.22 & 48.36 & 42.22 & 36.01 & 38.87
\\
 32  & 126\%& 33.60 & 51.67 & 40.72 & 59.50 & 69.61 & 64.16 & 49.04 & 51.08 & 50.04 & 45.00 & 34.92 & 39.33
\\
\hline
\end{tabular}
}
\end{center}

\caption{Classification accuracy (\%) of task incremental generalized zero-shot learning using RKR with different $K$ values.}

\label{tab:ablkvalue}
\end{table*}

\subsection{Implementation Details}
For the task incremental generalized zero-shot learning problem, we use the setting described in \cite{wei2020lzsl}. The authors in \cite{wei2020lzsl} propose a task incremental generalized zero-shot learning problem where each task is a separate dataset and show how a popular zero-shot learning approach, CADA-VAE \cite{schonfeld2019generalized} suffers from catastrophic forgetting in the image/visual feature encoder in this setting. We apply our approach to this setting and show that RKR achieves state-of-the-art results in this setting. In our approach, we apply the weight corrections and scaling transformations to the visual features encoder. We use $K=16$ to generate weight rectifications in this setting and report the average results of 5 runs for our method. Please refer to the supplementary materials for further details.

We apply RKR to the CADA-VAE framework, which contains only fully connected layers. SCM in CCLL only calibrates convolutional layer outputs (feature maps). Therefore, CCLL cannot be applied to CADA-VAE. We compare our method RKR with LZSL \cite{wei2020lzsl} and with the baseline methods proposed in \cite{wei2020lzsl} i.e., a) Sequential Fine-tuning (SFT): model is fine-tuned on new tasks sequentially, and the model parameters are initialized from the model trained on the previous task, b)  L1 regularization (L1): model weights are initialized with the weights of the model trained on the previous task, and the model is trained with an L1-regularization loss between the previous and current network weights, c) L2 regularization (L2): same as (b) but with L2-regularization loss, d) ``Base": model trained sequentially on all tasks without using any incremental learning methods or fine-tuning, e) ``Original": trains separate networks for each task.

\subsection{Results} \label{sec:zerores}

\begin{table*}[!htb]
\begin{center}
\scalebox{0.8}{
\begin{tabular}{ c| ccc | ccc | ccc | ccc }
\hline
 Initialization &  \multicolumn{3}{c}{aPY}  &  \multicolumn{3}{c}{AWA1}  & \multicolumn{3}{c}{CUB}  & \multicolumn{3}{c}{SUN} \\
  
\cline{2-13}
  & U   & S   & H   & U   & S   & H
 & U   & S   & H  & U   & S   & H \\
 \hline
 Random & 33.39 & 51.34 & 40.46 & 54.11 & 67.13 & 59.92 & 37.84 & 44.25 & 40.79 & 36.11 & 28.29 & 31.73
\\
 Previous & 33.39 & 51.34 & 40.46 & 58.79 & 69.36 & 63.64 & 47.52 & 49.22 & 48.36 & 42.22 & 36.01 & 38.87
\\
\hline
\end{tabular}
}
\end{center}

\caption{Classification accuracy (\%) of task incremental generalized zero-shot learning using our proposed RKR with different types of initialization: 1) random 2) from previous task.}
\label{tab:ablforward}

\end{table*}

Table \ref{tab:mainres} compares the performance of our method with the baselines, and LZSL \cite{wei2020lzsl} using the three evaluation metrics: unseen average class accuracy (U), seen average class accuracy (S), and harmonic mean of the two (H). The sequence of tasks/datasets is aPY, AWA1, CUB, and SUN, for a fair comparison with the other methods.

Table \ref{tab:mainres} also reports the total memory required by each method for the four tasks. 
LZSL requires 200\% memory for the image feature encoder as it stores the image features encoder trained on the previous task to calculate the knowledge distillation loss. The L1 and L2 baselines also require 200\% memory as they store the image features encoder trained on the previous task to calculate the L1/L2 loss between the weights of the two encoders. The ``Original" model trains four separate networks for the four tasks and requires 400\% total memory. Our method RKR requires around 3.28\% additional parameters for each task. Therefore, on four tasks, RKR requires a total of about 113\%  memory for the image features encoder.

The ``Base" model performs extremely badly on the first three tasks and manifests a clear case of catastrophic forgetting. SFT performs better than the ``Base" model since it fine-tunes the model on the new task. However, its performance starts dropping for the older tasks as it learns new tasks. The forgetting is lower in  SFT but is still substantial. We observe similar forgetting for the L1 and L2 baselines. Our method RKR significantly outperforms LZSL \cite{wei2020lzsl} as well as all the baseline methods. Specifically, RKR outperforms the state-of-the-art method LZSL by absolute margins of 5.65\%, 6.91\%, 6.33\%, and 2.53\% for the aPY, AWA1, CUB, and SUN datasets, respectively. We also compare the average H values across the four datasets. The average H values are 10.2\%, 36.73\%, 38.03\%, 36.73\% and 42.48\% for base, SFT, L1, L2 and LZSL \cite{wei2020lzsl} respectively. The average H value for RKR is 47.83\%, and that of the ``Original" model is 49.82\%. Therefore, RKR is significantly closer to the ``Original" model as compared to LZSL.

\textbf{Different First Task:} Table \ref{tab:ablseq} contains the results for different sequences of tasks/datasets having different first dataset. The H values for the AWA1 dataset with the first dataset as aPY, CUB, and SUN are 63.64\%, 62.24\%, and 62.46\%. Considering the fact that aPY, CUB, and SUN have a large variation in the number of classes (aPY = 32, CUB = 200, SUN = 717), this variation in the result is minor. We observe the same pattern for the other three tasks with different first tasks. 

\textbf{Significance of Components:} From Table \ref{tab:ablcomp} we observe that without weight rectifications, RKR performs lower by absolute margins of 6.8\%, 9.28\%, and 6.24\% for the AWA1, CUB, and SUN datasets, respectively. Similarly, without scaling, RKR performs lower by absolute margins of 3.71\%, 4.57\%, and 3.59\% for the AWA1, CUB, and SUN datasets, respectively.  Therefore, both weight rectifications and scaling factors are vital in this setting.

\textbf{Value of K:} Table \ref{tab:ablkvalue} reports performances of RKR with different values of $K$. $K=16$ performs close to $K=32$ for most of the datasets but requires significantly less total memory, i.e., 113\% vs. 126\%. Therefore, we choose $K=16$ for all our experiments in this setting, which significantly outperforms the state-of-the-art method.

\textbf{Forward Knowledge Transfer}\label{sec:ablforward}
In RKR, when a new task becomes available for training, we initialize the RG and SFG parameters from the previous task. We also experiment with training these parameters from scratch for each task. Table \ref{tab:ablforward} reports the performance of our method RKR with the two types of initialization for the RG and SFG parameters. When we initialize these parameters from scratch, the model performs lower by absolute margins of 3.72\%, 7.57\%, and 7.14\% from the other case, for AWA1, CUB, and SUN datasets, respectively. Therefore, forward transfer of knowledge takes place in RKR.

\section{Conclusion}

We propose a novel approach called Rectification-based Knowledge Retention (RKR) for the task incremental learning problem. RKR learns rectifications to adapt the network weights and intermediate outputs for every new task. We empirically show that our approach significantly outperforms the state-of-the-art methods for task incremental learning problem in both the zero-shot and non zero-shot settings. 
{\small
\bibliographystyle{ieee_fullname}
\bibliography{egbib}
}

\section{Supplementary Material}
\begin{figure*}[h]
    \centering
     \includegraphics[width=0.9\textwidth]{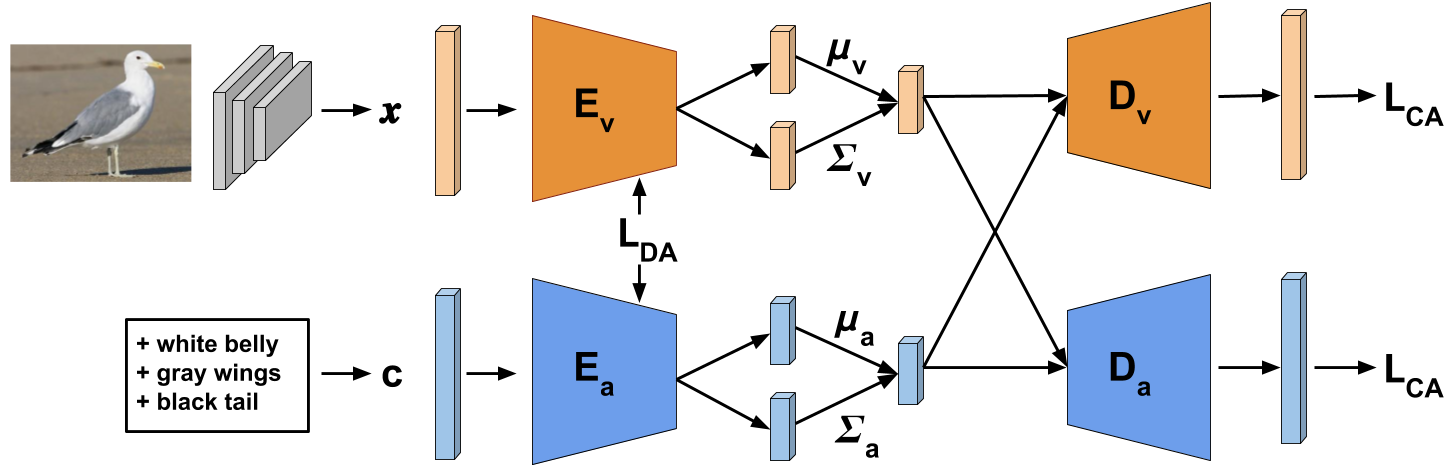}

    \caption{Illustration of CADA-framework.}

    \label{fig:cadaorig}
    
\end{figure*}
\subsection{Task Incremental Learning (Generalized Zero-Shot Setting)}
\begin{figure*}[h]
    \centering
     \includegraphics[width=0.9\textwidth]{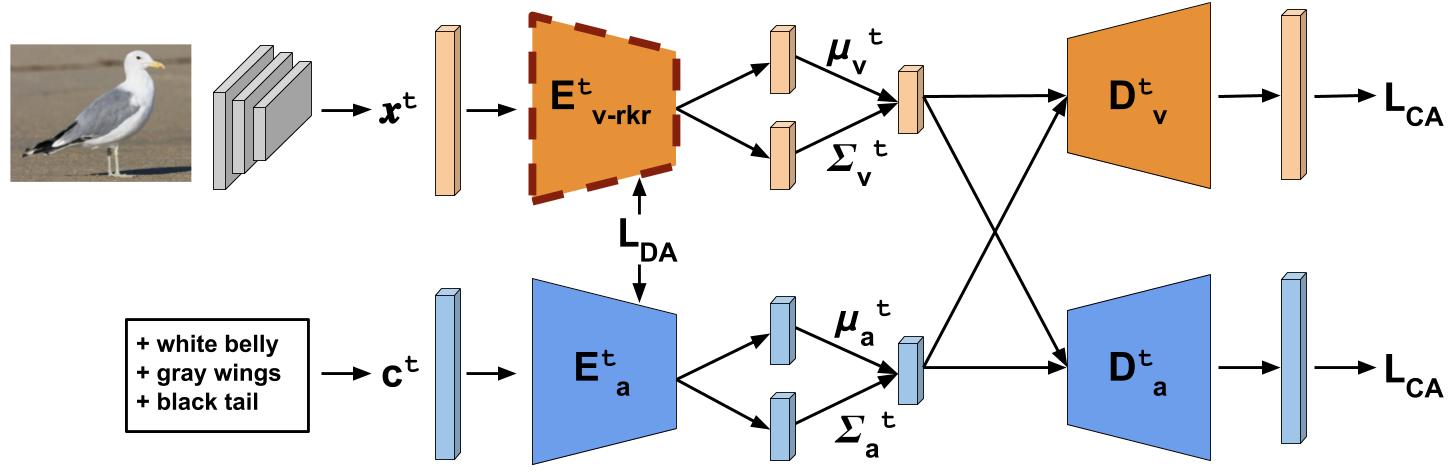}

    \caption{Illustration of CADA-framework with our proposed RKR image/visual features encoder ($E^t_{v\texttt{-}rkr}$). The framework consists of a pre-trained network for extracting image features and two variational autoencoders, one for the image/visual features and the other for the class/attribute embeddings. $E^t_{v\texttt{-}rkr}$ is our proposed RKR image/visual features encoder for task $t$ (shown with a dashed border), that adapts the network for task incremental learning. Image feature $x^t$, extracted by the pre-trained network from an image, is fed to $E_{v\texttt{-}rkr}$, which maps it to $\mu_v^t$ and $\Sigma_v^t$ in the latent space. The attribute encoder $E^t_{a}$ maps the attribute/class embedding $c^t$ of that image to $\mu_a^t$ and $\Sigma_a^t$ in the latent space. The network is trained on the standard VAE loss, cross-alignment loss $L_{CA}$ and distribution-alignment loss $L_{DA}$.
    }

    \label{fig:cada}
\end{figure*}
We use our approach RKR to work for the task incremental generalized zero-shot learning problem setting described in \cite{wei2020lzsl}. The authors in \cite{wei2020lzsl} experimentally show that CADA-VAE \cite{schonfeld2019generalized} suffers from catastrophic forgetting in the image/visual features encoder when trained in the task incremental generalized zero-shot learning setting. They propose LZSL to solve this problem. For a fair comparison with LZSL, we use the same setting, setup, and base architecture (CADA-VAE) as used in LZSL \cite{wei2020lzsl}.

\subsubsection{CADA-VAE}
In this section, we provide a brief overview of the CADA-VAE framework. For complete details, please refer to \cite{schonfeld2019generalized}. In the generalized zero-shot learning setting, there are S seen classes and U unseen classes, and we have labeled training examples for the seen classes only. The test images can be from both the seen and unseen classes. For each seen and unseen class, we have access to their class/attribute embeddings, which are generally vectors of hand-annotated continuous attributes or Word2Vec \cite{mikolov2013distributed} features. Zero-shot learning methods leverage the class/attribute embeddings to transfer information from seen classes to the unseen classes. Similar to most zero-shot learning methods, CADA-VAE operates on image features extracted by a pre-trained network (ResNet-101). 

The CADA-VAE framework consists of a variational autoencoder (VAE) for image/visual features ($E_v,D_v$) and a VAE for the class/attribute embeddings ($E_a,D_a$), each having an encoder and decoder (see Fig.~\ref{fig:cadaorig}). The two encoders project the image features and class embeddings to the common latent embedding space, respectively, and the decoders reconstruct the image features and class embeddings from their latent embeddings. Specifically, the image features encoder ($E_v$) maps the image features to $\mu_v$ and $\Sigma_v$ in the latent embedding space. Similarly, the class embeddings encoder ($E_a$) maps the class embeddings to $\mu_a$ and $\Sigma_a$. CADA-VAE learns a common latent embedding space for both the image/visual features and the class/attribute embeddings and brings the latent embeddings of the image features and class embeddings closer in the latent embedding space. It utilizes cross-alignment loss (CA) and distribution-alignment loss (DA) apart from the VAE loss to achieve this objective. Cross-alignment involves training the class embeddings decoder to generate corresponding class embedding from the latent features of the images of that class and vice-versa. Distribution-alignment involves training the encoders of the image features and class embeddings to minimize the Wasserstein distance between the Gaussian distributions of their latent embeddings. After training the VAEs, CADA-VAE uses the $\mu_a$ and $\Sigma_a$ of all the seen and unseen class embeddings to sample embeddings (using the reparametrization trick) for both the seen and unseen classes from the learned latent embedding space. It trains a classifier using these latent embeddings in order to classify the test images. At test time, the pre-trained network extracts image features from the test images. The image/visual features encoder ($E_v$) maps the test image features to the latent embedding space. The classifier then predicts the class for the test image latent embeddings.

\textbf{Task Incremental Generalized Zero-shot Learning:} The authors in \cite{wei2020lzsl} apply CADA-VAE to a task incremental generalized zero-shot learning setting where each task is a separate dataset and each task contain seen and unseen classes. After training the VAEs on a task $t$, embeddings can be sampled from the latent embedding space using the $\mu^t_a$ and $\Sigma^t_a$ of all the seen and unseen classes. These latent embeddings are used to train a classifier. The classifier will be able to predict the classes of the test image embeddings of that task produced by the visual features encoder ($E^t_v$).  However, when the network is trained on a new task $t+1$, the image/visual features encoder ($E^{t+1}_v$) weights will change. As a result, the test image features (input to $E^{t+1}_v$) for the test images from a previous task will get mapped to a different latent space (output of $E^{t+1}_v$) than the one obtained just after training the network on that task. Since the classifier will classify on the basis of the output of $E^{t+1}_v$, the predictions for the test images from the previous tasks will be incorrect. Therefore, the CADA-VAE performance suffers in the task incremental generalized zero-shot learning setting due to the catastrophic forgetting in the visual features encoder. The authors in \cite{wei2020lzsl} propose LZSL to tackle this problem by using selective parameter retraining and knowledge distillation to preserve previous task knowledge. 

\textbf{Applying RKR:} In order to prevent catastrophic forgetting, we apply RKR to the image/visual features encoder $E^t_v$, that only contains fully connected layers, to obtain $E^t_{v\texttt{-}rkr}$ (Fig. \ref{fig:cada}). Specifically, we use weight rectifications and scaling factors for each layer in the image/visual encoder to quickly adapt it to any task. We train the full network on the first task (base network). For every new task, we only learn weight rectifications and scaling factors for all network layers to adapt them to the new task. In the generalized zero-shot learning setting, we learn the weight rectifications and the scaling factors based on the seen classes of the given task and use them during testing for classifying both seen and unseen classes of that task. Therefore, during testing, the image features encoder will map the test image features for each task to the same embedding space as expected by the classifier.

\subsubsection{Datasets}
\begin{table}[htb]

\centering
\caption{Datasets used in the task incremental generalized zero-shot learning problem.}

\begin{tabular}{lccccc}
\hline
Dataset & Class Embedding & Images & \multicolumn{2}{c}{Classes}\\
 \cline{4-5}
 & Dimensions & & Seen & Unseen \\
  \hline
aPY & 64 & 15339 & 20 & 12\\
AWA1 & 85 & 30475 & 40 & 10\\
CUB & 312 & 11788 & 150 & 50\\
SUN & 102 & 14340 & 645 & 72\\
\hline

\end{tabular}

\label{tab:dataset}
\end{table}

For the task incremental generalized zero-shot learning problem, we experiment with the Attribute Pascal and Yahoo (aPY) \cite{farhadi2009describing}, Animals with Attributes 1 (AWA1) \cite{xian2018zero}, Caltech-UCSD-Birds 200-2011 (CUB) \cite{wah2011caltech}, and SUN Attribute dataset (SUN) \cite{patterson2012sun} datasets. Other details regarding the datasets are provided in Table \ref{tab:dataset}. For a fair comparison, we use the same sequence of training datasets given in \cite{wei2020lzsl}, which is aPY, AWA1, CUB, and SUN. However, we also report the results for three other cases with a different first dataset. We report the average per-class top-1 accuracy for the unseen classes (U), seen classes (S), and the harmonic mean of the two accuracies (H) for each dataset ($H = \frac{2\times U \times S}{U + S}$). Our objective is to achieve high H accuracy as it is not skewed towards either the seen or unseen classes. The results are obtained after the training has been completed on all the datasets.

\subsubsection{Implementation Details}

For our experiments, we extract the image features of 2048 dimensions from the final pooling layer of an ImageNet pre-trained ResNet-101. In the case of image features, the encoder and decoder of CADA-VAE have 1560 and 1660 hidden layer nodes, respectively. In the case of class embeddings, the encoder and decoder have 1450 and 660 hidden layer nodes, respectively. The latent embeddings are of size 64. For all the datasets, the model is trained for 100 epochs with batch size 50 using the Adam optimizer \cite{kingma2015adam}. CADA-VAE also uses a few hyper-parameters, i.e., $\delta$, $\gamma$, $\beta$. $\delta$ is increased from epoch 6 to epoch 22 by a rate of 0.54 per epoch, while $\gamma$ is increased from epoch 21 to 75 by 0.044 per epoch. The $\beta$ weight of the KL-divergence term is increased by a rate of 0.0026 per epoch up to epoch 90. A learning rate of 0.00015 is used for the VAEs, and a learning rate of 0.001 is used for the classifiers. L1 distance is used for the reconstruction error. These settings have been proposed in \cite{schonfeld2019generalized} and were also used in \cite{wei2020lzsl}. We report the average results of five runs for our method.

\subsection{Performance}
For task incremental learning, we perform five runs of every experiment for both the zero-shot and non zero-shot settings and report the average accuracy. The variations in our results are very low, e.g., for CIFAR-100 with ResNet18 and LeNet, the 95\% confidence interval for the final average session accuracy is 87.6 $\pm$ 0.467\% and 69.58 $\pm$ 0.55\% respectively.

\subsection{Hardware and Software Specifications}
We have performed all our experiments in PyTorch version 0.4.1 \cite{paszke2017automatic} and Python 3.0. For running our experiments, we have used a GeForce GTX 1080 Ti graphics processing unit.

\end{document}